\documentclass[conference]{IEEEtran}
\IEEEoverridecommandlockouts
\usepackage{cite}
\usepackage{amsmath,amssymb,amsfonts}
\usepackage{algorithmic}
\usepackage{graphicx}
\usepackage{textcomp}
\usepackage{subcaption}
\usepackage{xcolor}
\def\BibTeX{{\rm B\kern-.05em{\sc i\kern-.025em b}\kern-.08em
    T\kern-.1667em\lower.7ex\hbox{E}\kern-.125emX}}
\begin{document}

\title{Applications of Robots for COVID-19 Response\\}

\author{\IEEEauthorblockN{Robin R. Murphy}
\IEEEauthorblockA{\textit{Computer Science and Engineering} \\
\textit{Texas A\&M University}\\
College Station, TX \\
robin.r.murphy@tamu.edu}
\and
\IEEEauthorblockN{Vignesh Babu Manjunath Gandudi}
\IEEEauthorblockA{\textit{Computer Science and Engineering} \\
\textit{Texas A\&M University}\\
College Station, TX \\
vigneshbabu.gm@tamu.edu}
\and
\IEEEauthorblockN{Justin Adams}
\IEEEauthorblockA{\textit{Center for Disaster Risk Policy} \\
\textit{Florida State University}\\
Tallahassee, FL \\
jadams@em.fsu.edu}
}

\maketitle

\begin{abstract}
This paper reviews 262 reports appearing between March 27 and July 4, 2020, in the press, social media, and scientific literature describing 203 instances of actual use of 104 different models of ground and aerial robots for the COVID-19 response. The reports are organized by stakeholders and work domain into a novel
taxonomy of six application categories, reflecting major differences in work envelope, adoption strategy, and human-robot interaction constraints. 
Each application category is further divided into a total of 30 subcategories based on differences in mission. 
The largest number of reported instances were for public safety (74 out of 203) and clinical care (46), though robots for quality of life (27), continuity of work and education (22), laboratory and supply chain automation (21), and non-clinical care (13) were notable. 
Ground robots were used more frequently (119) than aerial systems (84), but unlike ground robots,  aerial applications
appeared to take advantage of existing general purpose platforms that were used for multiple applications and missions. 
Of the 104 models of robots, 82 were determined to be commercially available or already existed as a prototype, 11 were modifications to existing robots, 11 were built from scratch.
Teleoperation dominated the control style (105 instances), with the majority of those applications intentionally providing remote presence and thus not amenable to full autonomy in the future. Automation accounted for 74 instances and taskable agency forms of autonomy, 24.  The data suggests areas for further  research in autonomy, human-robot interaction, and adaptability. 

\end{abstract}


\section{Introduction}

The ongoing COVID-19 pandemic has already provided numerous instances of robotics being used to 
assist agencies with the detection and care of infected populations and prevent further transmission\cite{yang20}. Robots are also enabling industry,
educational institutions, and individuals to cope with the unique consequences of 
 sheltering-in-place restrictions and workplace absenteeism. 
 The purpose of this paper is to document the reported uses and perform a preliminary analysis. 
 This analysis is expected to provide guidance to policy makers and roboticists who
 are interested in increasing the immediate use of robots. It is also expected to provide a basis for further research and development of robots suitable for future pandemics and disasters. 
 
 The paper relies on 262 reports taken from weekly searches between March 27 and July 4, 2020 of press reports, social media including YouTube, Twitter, and Facebook using three search engines: google.com, social-search.com, and talkwalker.com. The description section and the comment section of the social media videos were also scraped manually to obtain additional links LinkedIn was searched manually. The searches used the following keywords and phrases: COVID, COVID19, COVID-19 robot, COVID19 Robots, COVID 19 Drone, COVID 19 UAS, COVID Drone, COVID UAS, ``COVID-19 and Robots", ``Use of Robots for COVID-19", ``Use of Robots for the present pandemic", and ``COVID-19 Robot uses".  The scientific literature was also searched. A preliminary content analysis was performed resulting in 203 instances of robots in actual use for COVID-19 related activities in 33 countries:
Belgium,
Chile,
China,
Cyprus,
Denmark,
Estonia,
France,
Germany,
Ghana,
Greece,
Honduras,
India,
Ireland,
Italy,
Japan,
Jordan,
Lithuania,
Mexico,
Netherlands,
Nigeria,
Philippines,
Rwanda,
Singapore,
South Korea,
Spain,
Sweden,
Taiwan,
Thailand,
Tunisia,
Turkey,
UAE,
US, and
the UK.

The reliance on press reports and social media may mean some applications are over- or under-represented; for example, a novel animal-like robot that is used in only a few 
locations for prototyping may receive more attention than commercially available disinfecting robots
used in hundreds of hospitals. Another limitation is that the search relied on queries in English; fortunately many non-English social media postings used English keywords such as ``robot" and contained pictures or videos. Despite the inherent incomplete nature of the reports, they indicate the general types of applications
and stakeholders (e.g., hospitals, private industry, individuals), which roboticists and policy makers can use to guide future development.

The article is organized as follows. Section~\ref{sec:taxonomy} presents a novel taxonomy of applications partitioned into six categories representing distinct use cases, stakeholders, and work envelopes. Each of the subsequent six sections describes the use cases for each category in more detail. Section~\ref{sec:findings} then discusses three sets of findings on how robots are used, what modality of robot (ground or air) is used and why, and the role of autonomy as a control scheme. The article concludes with topics for further research in autonomy, human-robot interaction, and adaptability.

\section{Methodology}
\label{sec:methodology}

The 262 reports from the weekly searches were copied into PDF and entered into a master spreadsheet, the master spreadsheet was divided into subsets of data for different analyses. This article considers only the subset of the data that captures actual robot use versus ethical concerns or general discussions.

Content analysis of the 262 reports produced 203 instances of a specific robot explicitly stated as being used for an application due to COVID-19 or a specific robot that would be used by a specific date for a COVID-19 related application (e.g., a purchase order had been placed for disinfection robots to cope with demand).  The instances did not include robots proposed for use or demonstrated in a laboratory; the robots had to be in use or in trials situated in the work environment. The report did not have to name the robot to be included if there was a video or image of the robot. When the robot was not specified, the media was inspected in order to try to recognize the robot or spot a logo.  In some cases, the report had links to  the manufacturer or showed logos; if the name of the company was identified, the company’s website, as well their LinkedIn page, as manually scraped in order to attempt to match the robot in the report.

Duplicate reports were combined into a single instance, as several robot stories were re-posted or covered by multiple news agencies. A single report could contain descriptions of multiple robots for different use cases, in which case the report produced multiple instances.  For example, there were two reports that documented the use of the LHF-connect robot for healthcare workers telepresence, patient and family socializing, as well as quarantine socializing; hence the reports produced 3 instances with one instance each in \textit{healthcare workers telepresence} and \textit{patient and family socializing} under \textit{Clinical Care} and \textit{quarantine socializing} under \textit{Non-Hospital} care respectively.

\section{Taxonomy}
\label{sec:taxonomy}

\begin{figure*}[htbp]
\centerline{\includegraphics[width=\textwidth]{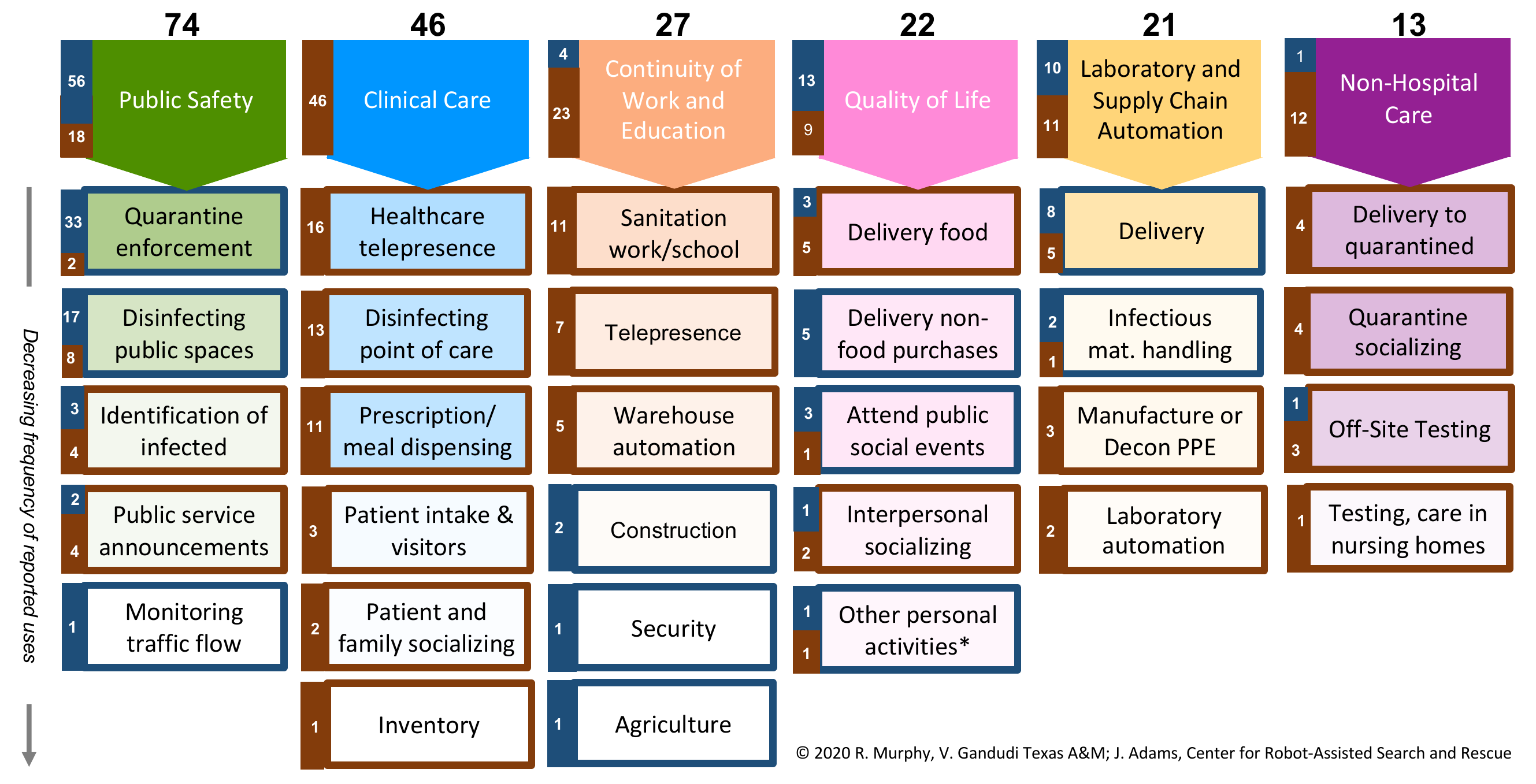}}
\caption{203 instances of robot use for COVID-19 applications organized into six categories and by modality, with brown representing ground robots and blue, aerial robots.}
\label{fig}
\end{figure*}

Figure 1 provides a graphical representation of the 203 instances grouped into
six applications by stakeholder and general work domain. The representation builds on a prior taxonomy introduced in \cite{conversation}. The advantage of grouping by stakeholder and work domain is that it makes it easier to consider three factors that influence the design of the robot: its work envelope (e.g., whether the work envelope is indoors or outdoors), how it is adopted (e.g., government agencies have different procurement procedures than private industry or individuals), and the human-robot interaction (e.g., are the users specialists or taken from the general population). 
The instances within a category are further subdivided
into subcategories to reflect differences in mission, work envelope, and socio-technical organization. Each application category is described in more detail in the following sections.
Figure 1 also shows the number of instances based on modality. At this time, only ground and aerial modalities have been reported. Of the 203 instances, 119 are ground and 84 are aerial. 

\section{Public Safety}

\textit{Public Safety} was the largest set of uses (74), with 56 instances using SUAS and 18 ground robots. The category consists of applications by law enforcement, emergency medical, and public works departments that may be coordinated with public health officials. The applications impact
the general population in public spaces. These public spaces can be outdoors, either urban areas or recreational areas, or indoors, such as subways. Normally these public spaces would be populated with citizens, which would impose safety constraints, but it appeared from the reports that most public spaces were either empty due to shelter-in-place restrictions or there were sparse collections of citizens. The applications in \textit{Public Safety}
are subject to jurisdictional polices and regulations. Additional regulations, such as aviation
restrictions on small unmanned aerial systems, may be present. In some cases, aviation restrictions in some countries were likely waived to permit operation in urban areas.  The use of robots
provided greater, and more timely, task coverage, high efficiencies than manual operations, and protected government workers from exposure. 

The largest reported use within \textit{Public Safety} was for enforcement of quarantine restrictions, primarily where law enforcement officers patrolled with SUAS (33) or ground robots (2) to detect the presence of large groups. In some cases, the robots were equipped with speakers and flew or drove close enough for the officers to give verbal warnings. 

The second largest use case was for disinfecting public spaces (25). SUAS for agricultural spraying were adapted to spray disinfectant in urban areas and large indoor facilities, while foggers were mounted on bomb squad or police ground robots. 

Ground and aerial robots were also used to attempt to identify whether a person in a public space was infected with coronavirus using thermal sensing (7), though the accuracy of such methods are in doubt CITE. Robots were also used to broadcast public service announcements about COVID-19 and social distancing (6). In one case, a SUAS was used to monitor traffic flow \textbf{WHY?} 

\section{Clinical Care}

\textit{Clinical Care} was the second largest set of uses (46), all with  ground robots. The 
category represents applications related to the diagnosis and acute healthcare of patients with coronavirus, These were  generally in a large hospital, or affiliated clinic, setting, where the hospital was presumed to have a reliable wireless network and information technology support.  The applications impact
the healthcare workers and support staff (i.e., janitors) and small segments of the general
population (i.e., patients and family members). The area of application were inside hospitals, with the diversity in waiting areas, treatment rooms, patient rooms, but also included entrances and parking garages where healthcare workers might be exposed.  Applications
are subject to public health, public health insurers, and hospital-specific polices and regulations.
Some policies and regulations may apply to robots if they are being used as medical devices for direct healthcare. However, the larger impact on robotics is whether the costs of acquiring and using robots can be reimbursed by insurers; if the insurers do not permit robots, then hospitals have less financial incentive to use them. 
The use of robots protected healthcare workers by enabling them to work remotely, allowed them to delegate unskilled tasks such as meal delivery and disinfection, and to cope with surges in demand. 

The largest reported use with \textit{Clinical Care} was for healthcare telepresence (16). This set of uses included use of teleoperation by doctors and nurses to interact with patients for diagnosis and treatment. 2 of the instances showed physical interaction, mostly to press sensors on the patient. 3 instances were physically invasive and the result of prior research in medical robotics. The robots were either commercially available or adapted from commercially available bases.

The second largest reported use was for rapidly disinfecting the hospital or clinic (13). The majority of robots used UVC light to perform a gross disinfection followed by a human wiping down surfaces likely to have been missed. Several models of this type of robot were already commercially available and in use for preventing hospital acquired infections. 

The third largest use was for prescription and meal dispensing (11), whereby carts navigated autonomously through a hospital. As with disinfection, these types of robots were already commercially available but their visibility and use increased as a means of coping with the surge in patients and the need to free the healthcare workers to spend more time on direct and compassionate care. 

The three other applications of robots for clinical care related activities were: use of telepresence robots in processing the intake of patients and handling families, essentially protecting the receptionists and clerks, (3) and in enabling families to remote visit patients (2); and in automating inventory management for a hospital floor (1).

\section{Continuity of Work and Education}

\textit{Continuity of Work and Education} was the third largest set of uses (27), primarily
with ground robots (23) and 4 instances of SUAS. The 
category represents applications where private industry and education leverage robotics for their enterprises, The work domain varied widely, from agriculture to classrooms. 
The applications impact small segments of the population, and the area of application was varied. 
The use of robots protected the workforce by enabling them to work remotely, allowed 
companies to delegate or automate tasks in order to cope with surges in demand, especially
for warehouses, or to cope with workforce reductions due to illness. 

Within \textit{Continuity of Work and Education}, the largest use was for sanitation of the workplace or school (11), leveraging the disinfection robots developed for hospitals. Telepresence
(7) and warehouse (5) robots were also leveraged. Robots were also used as assistants in construction of new hospitals (2),  to automate facility security (1) and agriculture (1) given the loss of personnel to extended illness. 

\section{Quality of Life}

\textit{Quality of Life} was fourth (22), split between
SUAS (13) and some ground robots (9). The 
captures how individuals are using robot-based services or applying personal robotics for day-to-day activities such as shopping and maintaining social connections.  The work domain varied widely, from urban to rural. 
The applications involved small segments of the population, spanning the spectrum of human-robot interaction from no interaction (e.g., packages dropped off at the individual's residence) to immersive interaction (e.g., socializing through robots). 
 The area of application was varied, with delivery robots working in urban and suburban areas and robots used for socialization in public venues or in homes.  
The use of robots allowed individuals to maintain the quality of life while complying with quarantining or sheltering in place restrictions. 

The largest use within Quality of Life was delivery (13). Delivery robots provided either food (8) or non-food purchases (5), where food delivery tended to use small platforms for a mission of transporting items for a single family. SUAS were more common (8) than ground robots (5). Social activities (9) was almost as large as delivery and split between SUAS (5) and ground robots (4). The use of robots for socializing reflecting great creativity, with individuals using a SUAS to ask another person out on a date and to walk a dog.  

\section{Laboratory and Supply Chain Automation}

The fifth largest use of robots was for 
\textit{Laboratory and Supply Chain Automation} was fifth (21), split between ground robots (11) and SUAS (10). Robots in this category were employed by private companies using robotics to support healthcare operations, either in hospitals, non-hospital facilities, or testing laboratories. 
The uses occurred in a wide range of work envelopes, including flying through urban areas, navigating through hospitals, and operating in manufacturing work cells. The interactions were
generally limited to humans trained to use the robots. 
The use of robots served two functions. One was to decrease the time for COVID-19 testing by reducing the time
spent transporting samples between clinics and the diagnostic laboratories, reducing the time
for delivery of reagents, and automating the test processes. The other was to protect healthcare workers by reducing exposure to coronavirus, either by robots handling biohazards or by making or cleaning PPE.

The largest reported uses within the \textit{Laboratory and Supply Chain Automation} category was
for delivery (13) and transportation of infectious materials (3). Stationary robots were used for manufacturing or decontaminating PPE (3) or laboratory automation (2). SUAS dominated the two transportation-oriented use cases (10 out of 16), leveraging the platforms used for delivery of food and non-food purchases to individuals. 

\section{Non-Hospital Care}

The sixth largest use of robots was for 
\textit{Non-Hospital Care}  (13), and almost exclusively with ground robots (12).  Robots in this category cover uses for non-acute care or diagnostics in nursing homes, temporary quarantine facilities, in-home healthcare providers, and corporate clinics; note that these settings seem similar to clinical care but have different missions, standards, economics, and stakeholders from acute care of infected. 
The uses occurred primarily in indoor setting mirroring \textit{Clinic Care} and the instance of off-site
testing mirrored \textit{Clinical Care}. The robots were used by healthcare workers and the general population, especially an elderly population. 
The use of robots was primarily to protect the elderly from exposure to coronavirus or to enable public health officials to handle the surge in quarantined individuals being housed in hotels or camps. 

\section{Findings}
\label{sec:findings}

While a complete analysis of the data is beyond the scope of this article, the 
data gathered offers insights into how robots are used, what modality of robot (ground
or air) is used and why, and the role of autonomy as a control scheme. 

\subsection{Uses}

The data shows that ground and aerial robots have an notable impact beyond direct clinical care, as only 46 of the 203 instances (22.66\%) were for \textit{Clinical Care}. This suggests that robotics is reaching
a level of adoption and that industry and the public are accepting robots in the workplace,
social arenas, and the home. 
The largest reported use was for \textit{Public Safety} (74), as compared to \textit{Clinical Care} (46). This could reflect
the visibility of public safety applications which, by definition, appear in view of the public, though
the low cost and availability of consumer SUAS most likely contributed to agencies innovating
creative uses. 

Delivery or transportation uses were pervasive, appearing in \textit{Clinical Care}, \textit{Quality of Life}, \textit{Laboratory and Supply Chain Automation}, and \textit{Non-Hospital Care}. This is interesting because of the differences in how
robots are adopted and by whom and suggests that delivery can become a general purpose application. 
Delivery or transportation indoors relied on ground robots (18), while outdoor movements
saw more use of aerial platforms (17) than ground (13). 

\subsection{Modality}

The data suggests that while ground robots are used more frequently (119), SUAS are
also emerging as important (84). SUAS seem to have an advantage as many platforms were commercially available and are built for general use, especially to carry different payloads, while ground robots were specialized for the
application, especially disinfection, dispensing, and delivery. 

104 different models of robots were identified, of which 82 were determined to be commercially available or already existed as a prototype, 11 were modifications to existing robots, 11 were built from scratch, and in 52 instances the origin of the robot could not be determined. 
The three most commonly used robots were the DJI Mavic 2 Enterprise UAS Duo (11 instances), TEMI UGV (9), and MMC UAS (6).

The 82 models of robots that were reported as being commercially available were predominately ground robots (66 models), with 16 models of UAS.
The 66 models of commercially available ground robots were: 
Temi (9), 
Xenex Light strike germ zapping robot (5), 
Double Robotics (4), 
Chloe (2), 
CloudMinds Robot (2),
Keenon Robotics Peanut (2),
Pudu Tech Robot (2), 
Spot (2), 
Tru-D (2), 
Ubtech Robotics Robot (2),
6 River Systems (1), 
Ana’s “Newme” Robot (1), 
Ava Robotics (1), 
Avid Bots Neo (1), 
Beam Pro (1),
Beep's Navya (1), 
BrainCorp Robot (1), 
Breezy One (1), 
Disinfection robot by Seoul Digital Foundation and Korea Robot Promotion agency (1),
EVA Robot (1),
Geek+ (1), 
Germ Falcon (1), 
Guangzhou Gosuncn Robot Co robot (1),
Intelligent Sterilization Robot (1),
InTouch Vici (1),
Ivo (1),
Jingdong Logistics Intelligent Distribution Robot (1), 
Kindred SORT (1), 
Knightscope (1),
Lions Bot (1), 
Miso Robotics (1),
Modai (1), 
Moxi (1), 
Ninja, University of Bangkok (1),
Nuro R2 (1), 
Opentrons OT-2 liquid handling robot (1), 
Pepper (1), 
P-Guard (1),
Postmate (1), 
Promobot (1), 
Qianxi Robot Catering (1), 
REV-1 (1), 
Robot for Care (1),
Seit-UV (1),
Seoul Digital Foundation and the Korea Robot Industry Promotion Agency Waste Disposal Robot (1),
Shanghai TMiRob (1),
Siasun Robot (1),
Skytron UV Robot (1), 
Starship Robot (1),
Sunburst UV Bots (1), 
Tally (1), 
Thermal temperature detection Robot by Seoul Digital Foundation and Korea Robot Promotion agency (1), 
Tommy Robot (1), 
Unity Drive Innovation (1), 
UVD Robot (1), 
Vaporized Hydrogen Peroxide Robot (1), 
Vision Semicon (1), 
White Rhino Auto Company (1), 
Whiz (1), 
Xiangdi (1),
Xiaofan Robot (1),
Zenplace’s Robot (1), 
Zen Zoe Robot (1),
Zhiping (1), 
Zorabots (1), and
Zorabot James(1).
The 16 models of commercially available aerial robots were: 
DJI Mavic 2 Enterprise Duo (11), 
MMC (4), 
Alphabet's Drone Delivery (2), 
Flytrex (2), 
Manna Aero's drone (2), 
Terra Drone Corporation drone (2), 
Zipline (2), 
DJI Phantom (1),
DJI Mavic Pro (1),
Everdrone (1), 
HTX Robotics (1),
Solent Transport (1), and 
Threod drone (1).
Verge Aero (1),
Wing (1), and
XAG (1),

Of the 11 models of robots that were reported as being modified for the COVID-19 application, seven were aerial vehicles and four were ground robots.
The following were the 8 models of existing aerial robots that were modified: 
DJI MG-1P (2),
MMC (2),
Alpha Drone Technology (1),
Eagle Hawk (1).
Hexa XL and Agras MG-1 (1), 
Netra Pro (1), and
XAG (1).
The following were the four models of existing ground robots that were modified: 
AIS K9 (1),
Soochow University spraying robot (1), 
Turtle Bot Variant (1), and
Youibot (1).

\begin{figure}
     \centering
     \begin{subfigure}[b]{0.3\textwidth}
         \centering
         \includegraphics[width=\textwidth]{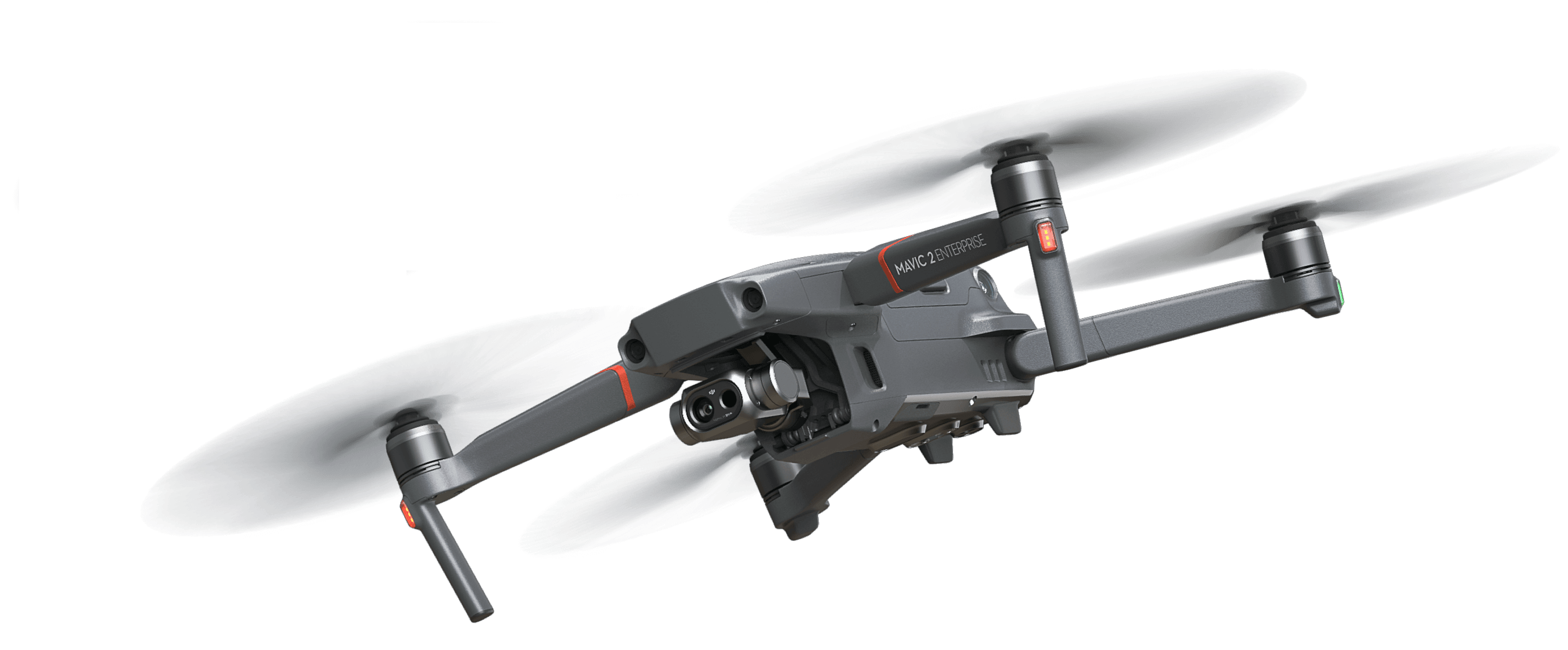}
         \caption{DJI Mavic 2 Enterprise}
         \label{fig:dji}
     \end{subfigure}
     \hfill
     \begin{subfigure}[b]{0.3\textwidth}
         \centering
         \includegraphics[width=\textwidth]{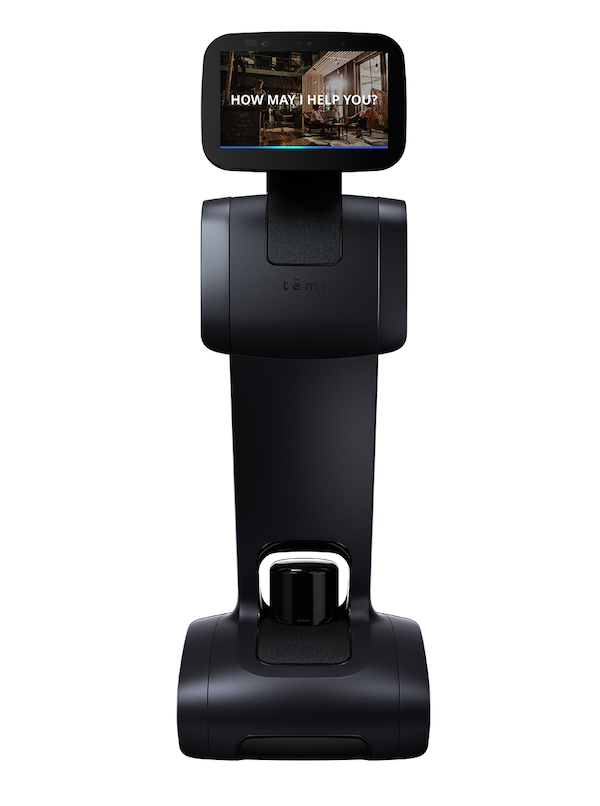}
         \caption{Robot TEMI}
         \label{fig:temi}
     \end{subfigure}
     \hfill
     \begin{subfigure}[b]{0.3\textwidth}
         \centering
         \includegraphics[width=\textwidth]{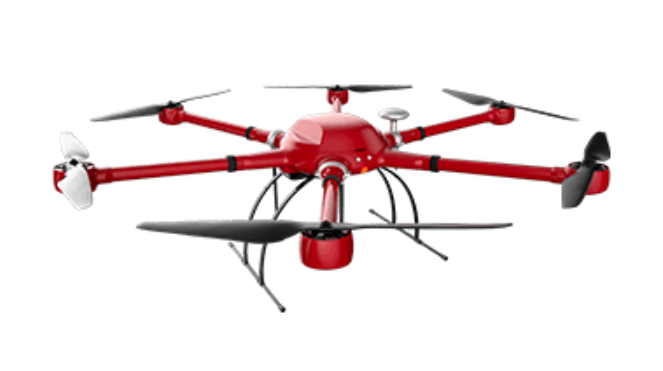}
         \caption{MMC}
         \label{fig:mmc}
     \end{subfigure}
        \caption{The three most commonly used robots reported: a) DJI Mavic 2 Enterprise, b) TEMI, and c) MMC (images from manufacturer's Website) }
        \label{fig:robots}
\end{figure}

The 11 models of ground robots built from scratch were: 
KARMI-Bot (3), 
LHF-Connect (3), 
Lego Bot to dispense sanitizer (2), 
Amazon’s disinfection robot (1), 
Cleantech-J1 (1), 
Guangzhou Institute of Respiratory Health and the Shenyang Institute of Automation’s throat swab robot (1), 
Pinto Robot (1), 
Tsinghua University’s robot (1), 
University of California Berkeley's Innovative Genomics Institute’s testing automation (1). 
University of Southern Denmark 3D printed robot (1), and
XDBOT (1 instance), 

The overall top three most commonly used models of robots 
were DJI SUAS models which accounted for 16 instances in 2 categories and 5 subcategories, with the Mavic 2 Enterprise Duo model used in 11 of the 16 instances, 
TEMI ground robot which accounted for 9 instances in 3 categories and 7 subcategories,
and 
MMC SUAS which accounted for 6 instances in 2 categories and 5 subcategories, Note that the most frequently used robots were used for different applications and missions, showing the value of general purpose platforms for disasters. See Figure~\ref{fig:robots}. DJI and MMC are made by Chinese companies, while TEMI is made by an Israeli company.  

For \textit{ Public Safety}, 15 of the 56 UAS instances deployed were commercially available, 9 instances were UAS with modifications, and the commercial maturity of UAS in 32 instances were unknown. DJI drones and Micromulti copter (MMC) and  were the most common. Different models of DJI drones accounted for 14 instances across 3 subcategories: Quarantine Enforcement, Disinfecting Public Spaces, and Public Service Announcements. MMC were used in 5 instances across 4 subcategories: Quarantine Enforcement, Disinfecting Public Spaces, Identification of Infected, and Monitoring Traffic Flow. 

Fifteen of the 18 instances with ground robots used commercially available UGVs, 1 used modified robots, and the commercial maturity of UGVs in 2 instances was unknown. 

For \textit{Clinical Care}, 32 of the 46 UGV instances deployed were commercially available, 2 instances were UGVs with modifications, 9 instances were UGVs built from scratch, and the commercial maturity of UGVs in 3 instances were unknown. Double Robotics Tele-presence robot and KARMI-Bot were the most common. Double Robotics Tele-presence robot accounted for 3 instances across 3 subcategories: Healthcare Workers Telepresence, Patient Intake and visitors, and Patient and Family Socializing. KARMI-Bot accounted for 3 instances across 3 subcategories: Healthcare Workers Telepresence, Disinfecting Point of care, and Prescription and Meal Dispensing.

For \textit{Continuity of Work and Education}, 16 of the 23 UGV instances deployed were commercially available, 1 instance was UGV with modifications, 5 instances were UGVs built from scratch, and the commercial maturity of UGVs in 1 instance was unknown. Xenex LightStrike Germ-Zapping Robot accounted for two instances across 1 subcategory: Sanitation at Work/School.  1 of the 4 instances with aerial vehicles used commercially available UAVs, and the commercial maturity of UAVs in 3 instances were unknown. XAG agricultural drone was the commercially available UAV deployed accounting for 1 instance in 1 subcategory: Agriculture.

For \textit{Quality of Life}, 10 of the 13 UAS instances deployed were commercially available, and the commercial maturity of UAS in 3 instances were unknown. DJI drones, Flytrex, Manna's Aero Drone and Alphabet's drone were the most common. DJI drones accounted for 2 instances across 2 subcategories: Interpersonal Socializing, and Other Personal Activities. Flytrex, Manna's Aero Drone and Alphabet's accounted for 2 instances each across 2 subcategories: Delivery Food, and Delivery Non-food Purchases. 6 of the 9 instances with ground robots used commercially available UGVs, and the commercial maturity of UGVs in 3 instances were unknown.

For \textit{Laboratory and Supply Chain Automation}, 9 of the 11 UGV instances deployed were commercially available, 1 instance was UGV built from scratch, and the commercial maturity of UGVs in 1 instance was unknown. 8 of the 10 UAS instances deployed were commercially available, and the commercial maturity of UAS in 2 instances were unknown. Terra Drone Corporations drone was the most common one. It accounted for 2 instances across 2 subcategories: Delivery, and Infectious Mat. Handling.

For \textit{Non-Hospital Care}, 10 of the 12 UGV instances deployed were commercially available, 1 instance was UGV built from scratch, and the commercial maturity of UGVs in 1 instance was unknown. Robot Temi was the most common one. It accounted for 6 instances across 4 subcategories: Delivery to Quarantined, Quarantine Socializing, Off-site testing and Testing-care in Nursing Homes. The commercial maturity the one UAS instance  was unknown.

\subsection{Control Scheme}

The 203 instances reflected three types of control: automation (74), teleoperation (105), and taskable agency (24). 
Taskable agency instances were often semi-autonomous, where only a portion of the mission was delegated
to the robot. As examples, consider i) a person wheels the robot into a room and then the robot performs the rest of the disinfection or ii) a meal delivery robot brings a meal to the door of a patient, but the nurse has to take the tray to the patient is too weak.  This incomplete
autonomy leads to hidden manpower costs because a person has to supervise or work with the robot, but the use of such robots can still provide a savings in time to complete tasks, 
reduced the effort and fatigue of workers, and protect workers from exposure.

Teleoperation was used in 17 out of 30 application subcategories:
quarantine enforcement,
disinfection of public spaces,
identification of infected,
public service announcements,
monitoring traffic flow,
healthcare telepresence,
patient intake and visitors,
patient and family socializing,
telepresence (work or school),
sanitation at work,
construction,
security
attend public social events,
interpersonal socializing,
other personal activities,
quarantined socializing, and 
off-site testing(Non-hospital care)
Of the 17 subcategories using teleoperation, 9 are remote presence applications where the human would wish to stay in the loop in order to apply their unique expertise or skills; have a rich, unscripted
interaction; and provide compassionate care:  
healthcare telepresence,
patient intake and visitors,
patient and family socializing,
telepresence (work or school),
attend public social events,
interpersonal socializing,
other personal activities,
quarantined socializing, and
off-site testing. These applications may benefit from increased intelligence
to make teleoperation easier but the human would not be eliminated. 
Automation or autonomy was used in 19 of the 30 application subcategories: 
disinfecting public spaces,
public service announcements,
disinfecting point of care,
prescription and meal dispensing,
patient intake and visitors,
inventory,
agriculture,
food delivery,
delivery non-food purchases,
sanitation of work or school,
warehouse automation,
interpersonal socializing,
delivery (laboratory-related),
manufacture or decontaminate PPE,
laboratory automation,
delivery to quarantined,
off-site testing,
testing,care in nursing homes, and
handling infectious materials. While autonomy and automation exists, the control schemes
may be subject to further improvements, such as faster navigation in more cluttered environments
with dynamic obstacles, and general reliability and safety. 
Six subcategories had both autonomous and teleoperation instances, suggesting that  autonomy is the ultimate goal
for a control scheme those applications: 
disinfecting public spaces,
public service announcements,
patient intake and visitors,
sanitation at work,
interpersonal socializing, and
off-site testing
These application subcategories appear to be ripe for advances in artificial intelligence. 

\section{Conclusions}

The  203 instances of robots being used explicitly to cope with the COVID-19 pandemic indicate that robots are being broadly used in all aspects of work and life, not just  for clinical care. The robots protect, not replace, healthcare and public safety workers by allowing them to work at a distance as well as handle the surge in duties and missions. The robots being used by businesses and individuals enable individuals to continue work and education, reduce the productivity impact of ill workers, protect at-risk populations in assisted-living facilities, and maintain social relationships. 

The data suggests areas for further  research in autonomy, human-robot interaction, and adaptability. 
While seven of the applications appear to be remote presence (LIST HERE) and thus would never eliminate the 
human, there are opportunities for increased intelligence to reduce the cognitive and training demands on the users. Likewise the autonomous uses were
still relatively limited. The instances suggest that more work is needed in no particular order:
\begin{itemize}
    \item navigation and coverage of public spaces- large, irregular, partial known with human presumably absent,
\item navigation and coverage of cluttered indoor spaces where humans may be present,
\item manipulation, both coarse (grasping for delivery) and dexterous (for patient care, handling instrumentation, retrieving inventory, etc.), 
\item physical HRI (especially  intrusive medical testing), and 
\item computer vision for manipulation, inventory, recognizing individuals, etc. 
\end{itemize}
Improved human-robot interaction is important, especially for i) interfaces for
teleoperation by incidental users and the general population to accomplish mundane task and 
ii) social interaction. 
The data also suggests that more research and development is needed for adaptable ground
robots. General purpose small unmanned aerial vehicles were rapidly re-purposed for new uses quarantine enforcement, disinfection, construction, and social interactions but ground robots showed less innovation. This analysis did not consider the attributes that lead stakeholders to adopt robotic innovations and there does not appear to be a model of responsible robotics innovation to guide reactive research and development for disasters; additional research on these topics is also warranted. 

The dataset is available upon request and data is updated and summaries are posted periodically to roboticsForInfectiousDiseases.org.

 \section*{Acknowledgment}
This work was supported by the National Science Foundation under Grant No. IIS-2032729; the opinions, findings,and conclusions or recommendations expressed are those of the author(s) and do not necessarily reflect the views of the National Science Foundation. It was conducted through the Center for Robot-Assisted Search and Rescue, a not for profit 501(c)(3) organization. The authors would like to thank Dr. Hao Su for his help in identifying occurrences in the scientific literature.


\bibliographystyle{./IEEEtran}
\bibliography{./covid.bib}

\end{document}